# Identifying Risk of Opioid Use Disorder for Patients Taking Opioid Medications with Deep Learning

Xinyu Dong, M.S.[1], Jianyuan Deng, M.Phil.[2], Sina Rashidian M.S.[1], Kayley Abell-Hart, B.S.[2], Wei Hou, Ph.D.[3], Richard N Rosenthal M.D.[4], Mary Saltz, M.D.[2], Joel Saltz, M.D., Ph.D.[2], Fusheng Wang, Ph.D.[1,2]

[1] Department of Computer Science, Stony Brook University, Stony Brook, NY

[2] Department of Biomedical Informatics, Renaissance School of Medicine at Stony Brook University, Stony Brook, NY

[3] Department of Family, Population and Preventive Medicine, Renaissance School of Medicine at Stony Brook University, Stony Brook, NY

[4] Department of Psychiatry, Renaissance School of Medicine at Stony Brook University, Stony Brook, NY

**Corresponding Author:**

Fusheng Wang

Department of Biomedical Informatics

Department of Computer Science

Stony Brook University

2313D Computer Science

Stony Brook, NY 11794-8330

Tel: +1 (631)632-2594

Email: fusheng.wang@stonybrook.edu

**Word Count**: 3,912

**Abstract**

**Background:** The United States is experiencing an opioid epidemic, and there were more than 10 million opioid misusers aged 12 or older each year. Identifying patients at high risk of Opioid Use Disorder (OUD) can help to make early clinical interventions to reduce the risk of OUD.

**Objective:** Our goal is to predict OUD patients among opioid prescription users through analyzing electronic health records with machine learning and deep learning methods. This will help us to better understand the diagnoses of OUD, providing new insights on opioid epidemic.

**Methods:** Electronic health records of patients who have been prescribed with medications containing active opioid ingredients were extracted from Cerner's Health Facts database between January 1, 2008 and December 31, 2017. Long Short-Term Memory (LSTM) models were applied to predict opioid use disorder risk in the future based on recent five encounters, and compared to Logistic Regression, Random Forest, Decision Tree and Dense Neural Network. Prediction performance was assessed using F-1 score, precision, recall, and AUROC.

**Results:** Our temporal deep learning model provided promising prediction results which outperformed other methods, with a F1 score of 0.8023 and AUCROC of 0.9369. The model can identify OUD related medications and vital signs as important features for the prediction.

**Conclusions:** LSTM based temporal deep learning model is effective on predicting opioid use disorder using a patient's past history of electronic health records, with minimal domain knowledge. It has potential to improve clinical decision support for early intervention and prevention to combat the opioid epidemic.

**Introduction**

Opioid Use Disorder (OUD) is a physical or psychological reliance on opioids, a class of substances found in certain prescription pain medications and illegal drugs like heroin[1], which includes opioid abuse and opioid dependence. Misuse and abuse of opioids are responsible for the deaths of more than 130 Americans daily[2], making it a leading cause of accidental death in the United States. Prescription opioids have been increasingly used due to its effectiveness in treating chronic pain[3]. According to Han et al[4], 91.8 million (37.8%) civilian non-institutionalized adults in the US consumed prescription opioids in 2015. Among them, 11.5 million (4.7%) misused them and 1.9 million (0.8%) had a use disorder. Prescription OUDs death has risen from over 4,000 to over 16,000 in 2010, making it the fastest growing form of drug abuse and overdose deaths involving opioids[5]. OUDs among individuals using prescription opioids are now a significant public health concern.

Earlier intervention in the developmental trajectory of OUD has the potential to reduce significant impairment and morbidity, if not mortality. For example, reducing opioid dosage or suggesting alternative options for chronic pain management can potentially reduce the risks of opioid use disorder. CDC has also provided recommendations for chronic pain care on opioid prescription for primary care clinicians[6]. In reality, if the risk of OUD can be predicted for targeted patient groups, early interventions can be made.

With the availability of electronic health records (EHR) of patients, we can build predictive models on the history of EHR to predict the risk of OUD and provide explanation of top risk factors. Based on that, clinical decision support can be made[7]. EHR has been widely adopted with the introduction of HITECH Act of 2009[8]. Besides EHR data managed by healthcare providers, large scale EHR data are also made available through commercial EHR vendors for research purposes. For example, Cerner's Health Facts[9] is a large multi-institutional de-identified database derived from EHRs and administrative systems.

Traditional statistical and machine learning based models on OUD prediction have been discussed in previous work[10-13]. For example, Cox regression method is applied to extract most relevant features and then to build a multivariate regression model to fit those features to predict two-year risk of opioid overdose[14]. Ellis RJ's[15] studied Gini importance, effect size and Wilcoxon rank-sum test to measure the importance of different features, and then applied a random forest classifier to predict opioid dependence. Except for random forest, decision tree and logistic regression were also proposed in Wadekar's work[16] for OUD prediction using demographic information, socioeconomic, physical and psychological features, and it identifies that the first use of marijuana before the age of 18 years is the most dominated predictor. Lo-Ciganic applied GBM and dense neural network models to build models for opioid overdose prediction using a set of hand-crafted features, including demographic information, medical code information, aggregated features such as daily morphine milligram equivalents[17, 18].

Recently, deep learning methods are gaining popularity in EHR based predictive modeling. For instance, Rajkomar et al. performed a large scale deep learning-based study with high prediction accuracy using EHR data in multiple medical events prediction[19]. Another study employed a fully connected deep neural network to suggest candidates for palliative care[20]. Temporal oriented deep learning models were also applied to solve relevant problems, for instance, Recurrent Neural Network(RNN) was proposed in Che's work to classify opioid users into long term users, short term users and opioid dependent patients, using diagnosis, procedure and prescription information[21]. Another study explored the application of RNN for chronic disease prediction using medical notes[22]. Our recent work has applied fully connected networks for predicting diseases and improving coding[23, 24] and opioid overdose prediction[25].

In this paper, we propose a temporal machine learning based prediction model built upon LSTM for predicting OUD among patients who have been prescribed with opioids medications using the history EHR data. The temporal based model can better model the progression of diseases. It can also identify the most important features for such predictions. We took advantage of past medical history including diagnoses codes, procedures codes, laboratory results, medications, clinical events and demographic information of patients for training a prediction model. We also compared our method with traditional machine learning algorithms and dense neural networks. Our results demonstrated that with comprehensive EHR data, our temporal deep learning model provided highly promising prediction results. The highest F-1 score achieved was from the LSTM model (precision: 0.8184, recall: 0.7865, f1 score: 0.8023, AUC: 0.9369). We also discovered that the model can identify OUD related diagnoses and medications as important features for prediction.

**Methods**

*Data Source*

*Cerner's Health Facts Database.* This database includes de-identified EHR data from over 600 participating Cerner client hospitals and clinics in the United States. In addition to encounters, diagnoses, procedures and patients' demographics that are typically available in claims data, Health Facts also includes medication dosage and administration information, vital signs, laboratory test results, surgical case information, other clinical observations, and health systems attributes[26].

*Data Selection*

As patients with opioids prescription are the target cohort of this study, we extracted all the patients who have been prescribed with medications containing active opioid ingredients in their medical records. For retrieval of those ingredients, we used the Anatomical Therapeutic Chemical (ATC) level 3 code 'N02A' and categories description 'opioid' to retrieve all relevant active ingredients from DrugBank 5.1.4[27]. Selected opioid related ingredients include butorphanol, diamorphine, eluxadoline, oxycodone, oxymorphone, naloxone, tramadol, levacetylmethadol, pentazocine, hydromorphone, levorphanol, remifentanil, normethadone, opium, sufentanil, piritramide, tapentadol, morphine, codeine, dezocine, fentanyl, nalbuphine, meperidine, naltrexone, buprenorphine, methadone, hydrocodone, alfentanil, dihydrocodeine, diphenoxylate.

Following procedures from Moore [28], we selected a group of ICD9 and ICD10 codes to define opioid use disorder patients. The summary of the codes can be found in Multimedia Appendix 1. Patients with one or more of these codes were considered OUD patients. Opioid medications have proven successful in treatment of cancer pain[29]. Cancer patients may receive many more opioid prescriptions than other patients, which may lead the model to misclassify these patients as having OUD, so we removed all patients with cancer diagnosis. To identify patients with cancer, we used the ICD-9[30] and ICD-10 codes[31], the summary can be found in Multimedia Appendix 2.

The majority of OUD patients (91.08%) is in the age group between 18 and 66, besides the number of patients with the highest increase rate is at 18, and the highest decrease rate is at 66. To make positive and negative cases consistent and prevent potential bias on ages, we filtered both OUD and non-OUD patients based on their age of first exposure to opioid medications between 18 and 66. The bar plot describing the age distribution of first opioid medication exposure for OUD patients is shown in Figure 1.

After age filtering, there are 111,456 positive (OUD) patients, and 5,072,110 negative patients. In the feature matrix for OUD patients we put together all information from encounters prior to the first encounter having an OUD diagnosis code; while for non-OUD patients, we did a similar process for information from all encounters except the last one.

*Feature Selection*

Information useful to predict future opioid use disorder includes diagnosis codes, procedure codes, lab tests, medications, clinical events and demographic information.

*Diagnosis codes* specify diseases, symptoms, poisoning for patients. This history of diseases is critical information for predicting the future. In Health Facts, both ICD-9 and ICD-10 codes exist. We converted all ICD-9 codes to ICD-10 codes to avoid dispersion of predictability for each diagnosis feature[32] and used the first 3 digits of the ICD-10 codes. Compared to the detailed ICD codes with all digits, the first 3 digits of ICD codes describe the general type of injury or disease, which is better for prediction, and can reduce the dimensions of features to accelerate the training process.

*Medications* are recorded by National Drug Code (NDC) code in Health Facts, which give detailed labeler, product and package information. To make better use of the action mechanism of medications, we converted all NDC codes to ATC codes because ATC codes annotate active ingredients by the system/organ they act on and their therapeutic/pharmacological class. Moreover, by using ATC codes, the number of medication related features is significantly reduced yet still those features are effective and carrying vital information. ATC level 3 codes were chosen[33] to represent all medications. For each medication, the total medication quantity prescribed to each patient was taken as a feature for each medication. In addition to the total medication quantity, we also calculated the amount of opioid ingredients contained in each medication and converted it to morphine milligram equivalents (MME) as an aggregate feature[17, 34].

*Lab Tests* are procedures in which a health care provider takes a sample of a patient's blood, urine, other bodily fluid or body tissue to get information about the patient's health. The numeric values for each test are recorded in Health Facts as well as the description of the value indicating whether it is higher than normal values, lower than normal values or within the normal range. We recorded the number of values higher and lower than normal values that the patient received and the total number of lab tests that the patients received.

*Clinical events* are related symptoms, procedures, and personal situations that are not formally classified into any codes above, for instance, the pain level of patients, smoking history, height, weight, and travel information. Since 79.21% of hospitals in Health Facts have clinical event records, they can be helpful for most hospitals for prediction.

*Demographic information* includes age, gender and race/ethnicity, they are added to the feature space as well to improve the prediction.

We extracted 1,468 features (457 diagnosis features, 530 laboratory test features, 3 demographic features, 251 clinical events features, and 227 medications features) to predict future opioid use disorder based on the patient's EHR history. Features of low occurrence were removed to prevent overfitting; if a feature had a prevalence in less than 1% of all opioid use disorder patients, that feature was not included in the feature space.

Table 1. Summary of features.

| Category | Number of Features | Description |
|---|---|---|
| **Diagnosis** | 457 | ICD9 code converted to ICD10, and use the first 3 digits |
| **Medication** | 227 | Total quantity patient has taken for each medication |
| **Clinical Events** | 251 | |
| **Laboratory Test** | 530 | Number of high, low and normal values for each test |
| **Demographics** | 3 | Gender, Age, Race/Ethnicity |

*Feature Matrix Construction*

We used a binary representation for diagnosis codes to record its existence in a patient's history. For ages, we segmented them into multiple age groups: the first age group is 18-27, followed by every 10 years. Race/ethnicity were encoded by one-hot encoding, which is the most common coding scheme for categorical variables, it transforms a single variable with n distinct values to n binary variables indicating presence (1) or absence (0). Other features recorded in numeric values, such as medication dosage, clinical events of numeric values such as blood pressure, height, pain score, were assigned with numeric values in corresponding units in the database, such as height measured in cm, blood pressure measured in mm(Hg). Numeric values for each laboratory test are recorded in Health Facts with the description of the value indicating whether it is higher than, lower than, or within the normal range. If the patients didn't do any laboratory test in one encounter, we will record 0 for low and high value portion, 1 for normal value portion. In addition, for some clinical events, there is multiple values in one encounter, for example, body temperature may be tested multiple times in one visit. We recorded the highest, lowest and median value of the feature in that encounter. If there is missing value, we replaced them with median values for numeric features, while we will replace them with majority values for binary features.

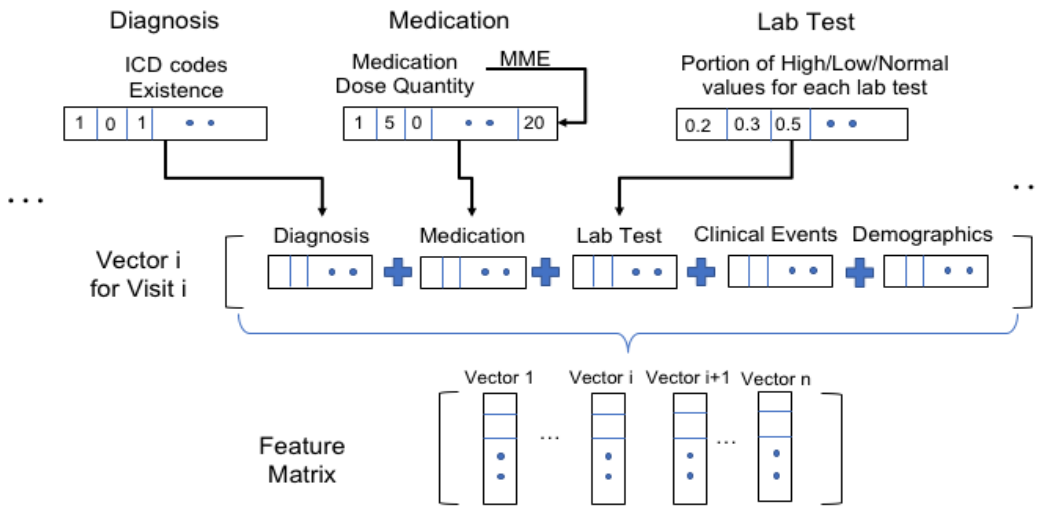

Figure 1. Structure of dense neural network model.

*Dense Neural Networks*

Deep neural networks have been proven effective in many healthcare prediction applications[19-22] due to the capability of handling large numbers of features, which is well suited for our problem. We implemented a fully connected neural network model. Our model is composed of six fully connected layers, each of first 5 layers has a dimension of 512, takes relu as the activation function and is followed by a dropout of 0.3. The last layer has a dimension of 8 and is connected to the output layer with binary cross-entropy loss function and Adam optimizer. Dropout layer will randomly drop out a portion of outputs from the previous fully connected layer at each training epoch to prevent overfitting. The framework of our neural network is illustrated in Figure 3.

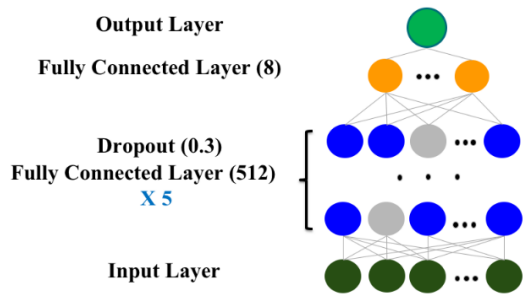

Figure 3. The structure of the dense neural network model.

*The LSTM Model*

A recurrent neural network (RNN) is a class of artificial neural networks where connections between nodes form a directed graph along a temporal sequence. This makes the RNN model well suited to make predictions based on time series data. Long Short-Term Memory (LSTM) networks are a modified version of recurrent neural networks, which makes it easier to store past data in memory. Also compared to RNN, it can solve the vanishing gradient problem in common RNN models, so we employed the LSTM model in our problem[35]. We implemented an LSTM based network composed of two LSTM layers of 512 units. The framework of our LSTM based neural network is illustrated in Figure 4. In our experiment, the implementation environment is the Python programming language (2.7). Traditional machine learning methods are implemented with Python Scikit-Learn package[36]. Deep learning is implemented with Python Tensorflow[37] and Python Keras[38]. Other used libraries include Python Numpy[39] and Python Pandas[40]. The training was performed on an NVIDIA Tesla V100 (16GB RAM).

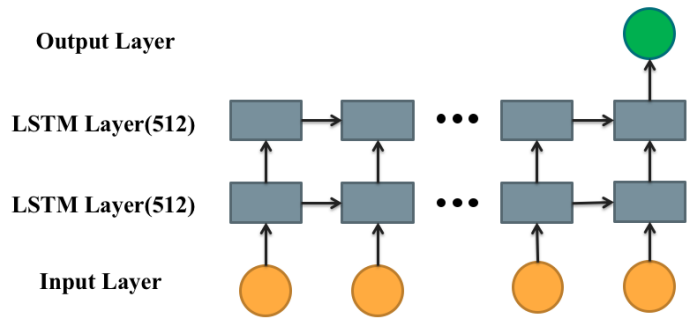

Figure 4. Structure of LSTM model.

**Results**

In our experiments, we randomly took 80% of patients as the training set and the rest as test set. For each method, we repeated the training process 10 times then to calculate an average value for each metric to assess model performance. In our experiments, we portioned the total cohort of negative patients into 10 parts of 507211 negative patients each. We trained each part with positive patients separately and randomly selected 80% as a training set and the remaining 20% as the test set. For each metric to evaluate a model's performance, we calculated the average for all 10 parts.

To comprehensively evaluate the models, we calculated all common metrics including precision, recall, F1 score and Area Under the ROC Curve (AUC score). As the dataset is imbalanced, AUC can be misleading [41]. Recall is a critical factor for the prediction models, since it means how many potential OUD patients we can identify in advance, but it can also be achieved by playing parameters with a lower precision since there is a tradeoff between recall and precision. Therefore, the F1 score as a measurement considering both precision and recall, is regarded as best aggregated assessment of the overall prediction performance. The results of each metric for each method are shown in Table 2. The best results for each metric category are highlighted in bold. Results demonstrated our intelligent models are capable of classifying opioid poisoning very well.

Table 2. Summary of results.

| Model | Precision | Recall | F1 score | AUCROC |
|---|---|---|---|---|
| **Random Forest** | **0.8565** | 0.6871 | 0.7545 | 0.9112 |
| **Decision Tree** | 0.7592 | 0.7281 | 0.7453 | 0.8823 |
| **Logistic Regression** | 0.7507 | 0.6020 | 0.6722 | 0.7933 |
| **Dense Neural Network** | 0.8019 | 0.7694 | 0.7855 | 0.9224 |
| **LSTM** | 0.8184 | **0.7865** | **0.8023** | **0.9369** |

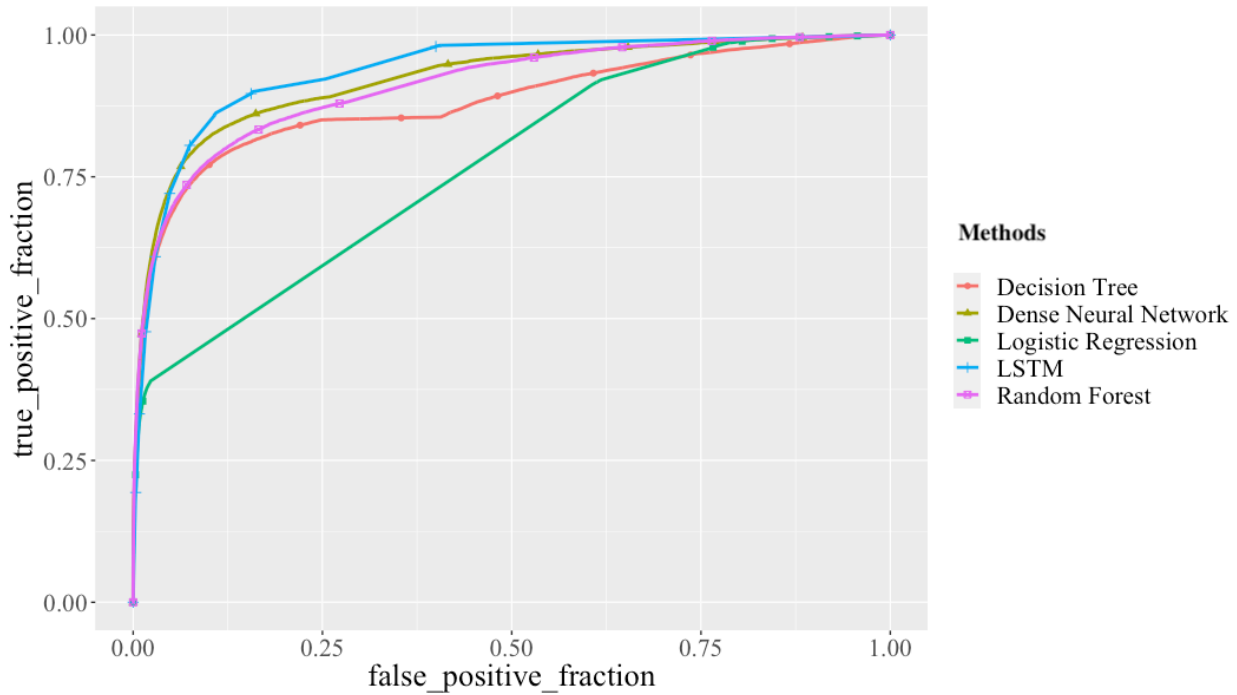

Figure 5. ROC Curve for All Methods.

We can achieve a F1 score of 0.8023 by the LSTM model. For precision, the best performance we can achieve is 0.8565 through Random Forest, and the LSTM model can also achieve a score of 0.8184, which indicates the accuracy when we identify a patient as OUD patient. For recall, we can achieve 0.7865 for the best case, indicating the ability to find all the OUD patients in the dataset. The best AUC score we can achieve is 0.9369 which indicates a good performance for clinical psychology applications with AUC values of greater than 0.747[42]. Figure 5 showed the ROC curves for all 5 methods.

*Missing Values*

There is often concern about the effect of missing value in prediction tasks. In our model, we used median imputation methods to deal with missing value problem. We are also interested in comparing median imputation with other common imputation methods including mean imputation, MICE and KNN. Since we already had LSTM model as the best method, we would compare the prediction by reperformance the experiment using LSTM with all those imputation methods. The result is shown in Table 3.

Table 3. Summary of LSTM prediction performance on different missing values.

| Method | Precision | Recall | F1 score | AUCROC |
|---|---|---|---|---|
| **Median** | 0.8184 | 0.7865 | 0.8023 | 0.9369 |
| **Mean** | 0.8183 | 0.7851 | 0.8014 | 0.8977 |
| **MICE** | 0.8128 | 0.8017 | 0.8072 | 0.9002 |
| **KNN** | 0.7984 | 0.8034 | 0.8009 | 0.8854 |

As we can see in the results, there is no significant difference among all imputation methods. Because the difference in prediction result is small, we conclude that these features do not make a tangible difference. Besides we can also find in previous research, there is no significant difference in results when comparing different imputation methods in noisy and large-space EHR data[23, 24].

*Top Features Analysis*

To support researchers or clinicians to exploit potential causes or trajectories of diseases, it is necessary to understand the importance of different features for prediction. Deep learning models are especially difficult to interpret especially because of their complex structures and millions of parameters. To evaluate the importance of features in deep learning models, we employed the permutation importance method. It measures the decrease of metrics scores by blinding a feature as the importance of that feature, the feature is more important when the score has bigger decrease[43]. We used the F-1 score as the measurement we want to calculate the decreases. This method can be used to all kinds of models including traditional machine learning models and deep learning methods. Python package of eli5 can be used to generate the importance[44]. Based on our best method LSTM based model, we generated top 50 important features in Figure 6 and Supplement Table 3.

Diagnosis and medication information dominate the top 50 important features. In particular for medications features, dose quantity of opioid medications (ATC Level 3 code N02A*) and MME are directly related to OUD. Besides, "N05B: Anxiolytics" is also a top feature. One explanation is that anxiety is common in patients with OUD. For top important diagnoses, many are related to pain, which are likely to be the cause of opioid use initiation for pain treatment either medical use or non-medical use. For instance, Dorsalgia (ICD10 codes M45.*) includes the chronic pain in back, pain not elsewhere classified (ICD10 codes G89.*), acute abdomen(ICD10 codes R10.*), and joint or tissue disorder(ICD 10 codes M25.* and M79.*) can also cause chronic pain. We can find that most of the top 50 important features are highly related to OUD, which indicates that our LSTM based model is capable of capturing relations between features and OUD.

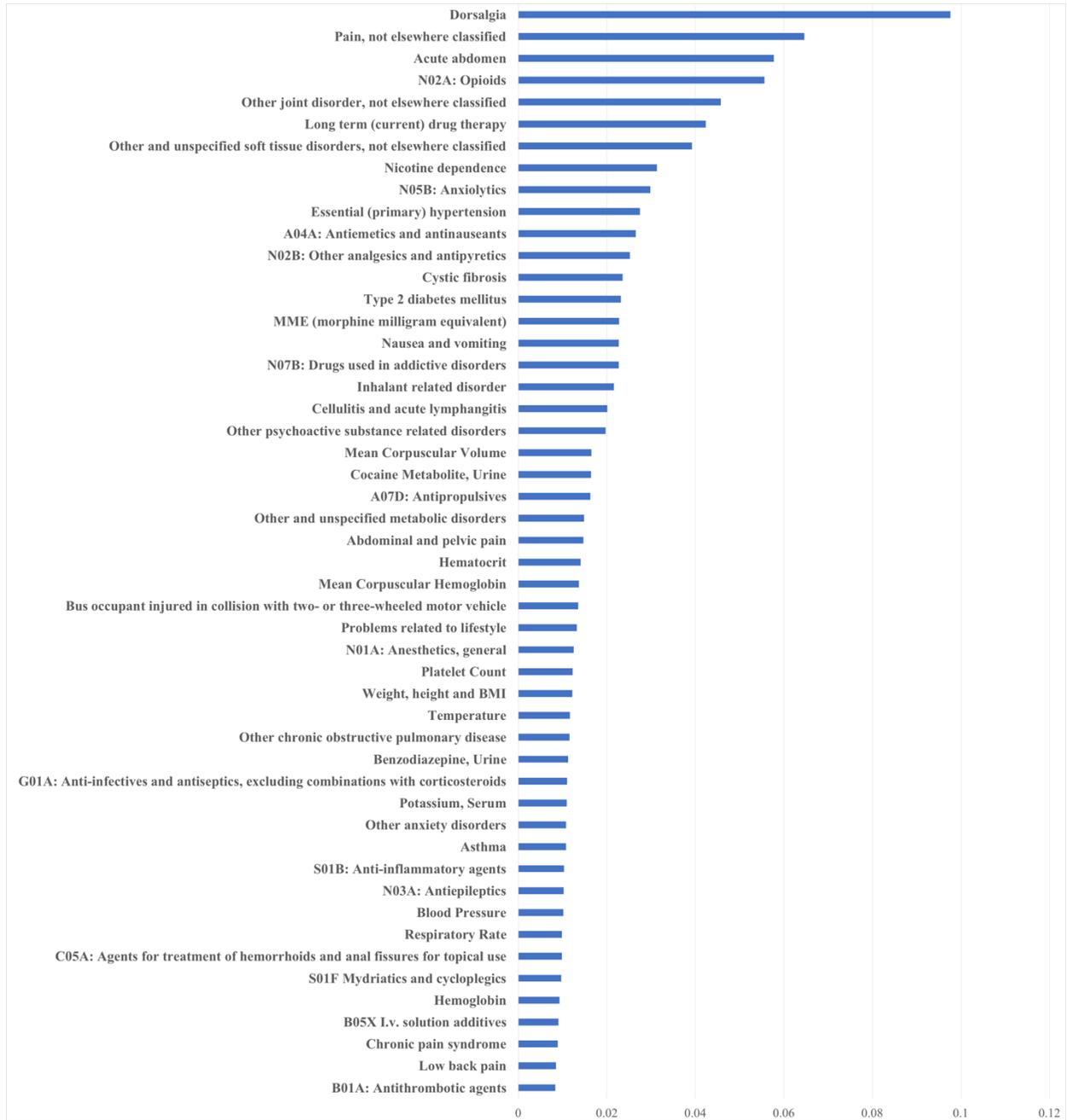

Figure 6. Top 50 Important Predictors for OUD Selected by LSTM.

**Discussion**

Data driven studies hold high potential for studying the problem of opioid epidemic in the U.S. With the wide availability of Electronic Health Records, predictive modeling provides a powerful approach to automatically predict the risks of opioid use disorder for patients who used prescription opioids.

LSTM model achieved a promising result with the best F1 score by LSTM model. Top important features generated from our model by permutation importance method also revealed interesting relationships. Chronic pain management and treatment of acute pain are among top factors leading to opioid use disorder.

*Comparison with previous work*

In Wadekar's work for OUD prediction, they applied random forest and other traditional methods on hand-crafted features[45]. Our AUC scores had a significant improvement from their work with AUC score of 0.8938. Lo-Ciganic applied Gradient Boosting Model (GBM) and dense neural network models to build models for opioid overdose prediction using a set of 268 hand-crafted features, they had an AUC score around 0.90, which is also inferior to our result. These methods used a limited set of features and lacked the modeling of temporal progression with state-of-the-art methods. Our approach can take advantage of as much information as possible to discover hidden relations.

*Benefits of the model*

Comparing top predictions with random forest, logistic regression, decision tree and dense neural network, LSTM predictions are notably more accurate. Compared to previous work, our study has several advantages. First, our study employed more comprehensive information, previous studies only include a limited set of features of clinical information, either diagnosis, medications or demographics, which are all included in our model. Secondly, for feature engineering, many previous works are based on professional knowledge, they require clinical knowledge to make hand-crafted features for the prediction. While in our model, it is depended on the deep learning model's ability to process large scale of data with minial domain knowledge needed. Thirdly, while a disease is often a progressive process, traditional methods are not built with such characteristics. Our LSTM based model can embody such characteristic when processing data. Fourthly, compared to basic RNN models, LSTM model has advantage for its ability to deal with vanishing gradient problems and insensitivity to gap length [35].

*Clinical Significance*

Understanding what is attributable risk for the development of OUD is critical for understanding how to construct a prevention response that may curtail the progression from incidental non-medical use of prescription opioids to habitual use, and compulsive use in OUD. Pain is reported as the most common motivation for OUD in many studies of adults. They would use opioid medications to treat pain, while a lot may use non-prescribed opioid. Those non-medical opioids use history is hard to track, but overuse of opioids can result in some symptoms. Our study is capable of finding those potential relations from including those features indicating the related symptoms. Except for the pain, OUD is also investigated to be correlated to factors, temperamental and personality traits, trauma and stressful life events, mental disorders. Our model can also help to discover the relations between OUD and those factors.

*Limitations*

The population of the study is derived from patients based on structured EHR records, and it does not capture on-prescribed opioids or unrecorded use disorder. Clinical notes are not included in the dataset, which may help to provide additional knowledge to improve the model.

**Conclusion**

The opioid epidemic has become a national emergency for public health in the United States. Predicting risk of opioid use disorder for patients taking prescription opioids can provide targeted, focused early interventions for smarter and safer clinical decision support. Our LSTM based deep learning predictive model of OUD using the history of EHR data demonstrate promising results. The temporal deep learning model is capable of identifying patients who will develop into OUD in the future and can provide critical factors on the risk. Our approach can potentially reduce OUD through earlier intervention in the developmental trajectory.

**Acknowledgements**

**Funding**

This work was funded partially by the Stony Brook University OVPR Seed Grant 1158484-63845-6.

**Conflict of Interest Disclosures**

The authors do not have any conflicts of interest to disclose.